\documentclass[preprint,10pt]{elsarticle}

\usepackage{textcomp,booktabs}
\usepackage[usenames,dvipsnames]{color}
\usepackage{colortbl}
\definecolor{mygray}{gray}{.9}
\definecolor{mypink}{rgb}{.99,.91,.95}
\definecolor{mycyan}{cmyk}{.3,0,0,0}
\usepackage{amssymb}
\usepackage{graphicx}
\usepackage{booktabs}
\usepackage{tabularx}
\usepackage{array}
\usepackage{lscape}
\usepackage{longtable}
\usepackage{multirow}
\usepackage{amsmath}
\usepackage{colortbl}
\usepackage{epsfig}
\usepackage{epsf}
\usepackage{subfigure}
\usepackage{rotating}
\usepackage{pdflscape}
\usepackage{lscape}
\usepackage{indentfirst}
\usepackage{wrapfig}
\usepackage{enumerate}
\usepackage{amsfonts,amssymb}
\usepackage{dsfont}
\usepackage{CJK}
\usepackage[linesnumbered,ruled]{algorithm2e}
\usepackage[amsmath,thmmarks]{ntheorem}
\biboptions{numbers,sort&compress}

\newcommand{\PreserveBackslash}[1]{\let\temp=\\#1\let\\=\temp}
\newcolumntype{C}[1]{>{\PreserveBackslash\centering}p{#1}}
\newcolumntype{R}[1]{>{\PreserveBackslash\raggedleft}p{#1}}
\newcolumntype{L}[1]{>{\PreserveBackslash\raggedright}p{#1}}
\newtheorem{definition}{Definition}[section]

\begin{document}

\begin{frontmatter}

\title{Generalized Belief Function: A new concept for uncertainty modelling and processing}

\author[address1]{Fuyuan Xiao\corref{label1}}
\ead{xiaofuyaun@swu.edu.cn}
\address[address1]{School of Computer and Information Science, Southwest University, Chongqing, 400715, China}
\cortext[label1]{Corresponding author at: School of Computer and Information Science, Southwest University, No.2 Tiansheng Road, BeiBei District, Chongqing, 400715, China. 
}

\begin{abstract}
In this paper, we generalize the belief function on complex plane from another point of view.
We first propose a new concept of complex mass function based on the complex number, called complex basic belief assignment, which is a generalization of the traditional mass function in DSE theory.
On the basis of the definition of complex mass function, the belief function and plausibility function are generalized.
In particular, when the complex mass function is degenerated from complex numbers to real numbers, the generalized belief and plausibility functions degenerate into the traditional belief and plausibility functions in DSE theory, respectively.
\end{abstract}


\end{frontmatter}

\section{Introduction}\label{Introduction}
In the past few years, to make the traditional Dempster--Shafer evidence (DSE) theory more practical, many researchers have extended the belief structure and Dempster's combination rule in different aspects.
In this paper, we focus on the generalization of belief function on the view of a new perspective, i.e., complex plane.
In particular, we first propose a new concept of complex mass function, called complex basic belief assignment based on the complex number.
The newly defined complex mass function is a generalization of the traditional mass function in DSE theory.
On the basis of complex mass function, the belief function and plausibility function are generalized.
Especially, when the complex mass function is degenerated from complex numbers to real numbers, the generalized belief and plausibility functions degenerate into the traditional belief and plausibility functions, respectively.

\newtheorem{Remark}{Remark}
\newtheorem{Property}{Property}
\newtheorem{Theorem}{Theorem}
\newtheorem*{Proof}{Proof}

\section{Preliminaries}\label{Preliminaries}

\subsection{Complex number~\cite{ablowitz2003complex}}\label{Complexnumber}
A complex number $z$ is defined as an ordered pair of real numbers
\begin{equation}\label{eq_complexnumber}
z = x + yi,
\end{equation}
where $x$ and $y$ are real numbers and $i$ is the imaginary unit, satisfying $i^2 = -1$.
This is called the ``rectangular'' form or ``Cartesian'' form.

It can also expressed in polar form, denoted by
\begin{equation}\label{eq_complexnumber}
z=r e^{i \theta},
\end{equation}
where $r > 0$ represents the modulus or magnitude of the complex number $z$ and $\theta$ represents the angle or phase of the complex number $z$.

By using the Euler's relation,
\begin{equation}\label{eq_Euler'srelation}
e^{i \theta} = \cos(\theta) +i \sin(\theta),
\end{equation}
the modulus or magnitude and angle or phase of the complex number can be expressed as
\begin{equation}\label{eq_magnitudeandphase}
r=\sqrt{x^2+y^2}, \text{ and } \theta = \arctan(\frac{y}{x}) = \tan^{-1}(\frac{y}{x}),
\end{equation}
where $x = r \cos(\theta)$ and $y = r \sin(\theta)$.

The square of the absolute value is defined by
\begin{equation}\label{eq_squaremagnitude}
|z|^2=z\bar{z}=x^2+y^2,
\end{equation}
where $\bar{z}$ is the complex conjugate of $z$, i.e., $\bar{z}=x - yi$.

These relationships can be then obtained as
\begin{equation}\label{eq_relationship}
r=|z|, \text{ and } \theta = \angle z,
\end{equation}
where if $z$ is a real number (i.e., $y = 0$), then $r = |x|$.

The arithmetic of complex numbers is defined as follows.

Give two complex numbers $z_1=x_1 + y_1i$ and $z_2=x_2 + y_2i$,

\begin{itemize}
\item
The addition is defined by
\begin{equation}\label{eq_addition}
z_1+z_2=(x_1 + y_1i)+(x_2 + y_2i)=(x_1+x_2)+(y_1+y_2)i.
\end{equation}

\item
The subtraction is defined by
\begin{equation}\label{eq_subtraction}
z_1-z_2=(x_1 + y_1i)-(x_2 + y_2i)=(x_1-x_2)+(y_1-y_2)i.
\end{equation}

\item
The multiplication is defined by
\begin{equation}\label{eq_multiplication}
(x_1 + y_1i)(x_2 + y_2i)=(x_1x_2-y_1y_2)+(x_1y_2+x_2y_1)i.
\end{equation}

\end{itemize}

\subsection{Belief function theory~\cite{Dempster1967Upper,shafer1976mathematical}}\label{BBA}

\begin{definition}(Frame of discernment)

Let $\Omega$ be a set of mutually exclusive and collective non-empty events, defined by
\begin{equation}\label{eq_Frameofdiscernment1}
 \Omega = \{F_{1}, F_{2}, \ldots, F_{i}, \ldots, F_{N}\},
\end{equation}
where $\Omega$ is a frame of discernment.

The power set of $\Omega$ is denoted as $2^{\Omega}$,
\begin{equation}\label{eq_Frameofdiscernment2}
\begin{aligned}
 2^{\Omega} = \{\emptyset, \{F_{1}\}, \{F_{2}\}, \ldots, \{F_{N}\}, \{F_{1}, F_{2}\}, \ldots, \{F_{1}, \\
 F_{2}, \ldots, F_{i}\}, \ldots, \Omega\},
\end{aligned}
\end{equation}
where $\emptyset$ represents an empty set.

If $A \in 2^{\Omega}$, $A$ is called a proposition.
\end{definition}

\begin{definition}(Mass function)

A mass function $m$ in the frame of discernment $\Omega$ can be described as a mapping from $2^{\Omega}$ to [0, 1], defined as
\begin{equation}\label{eq_Massfunction1}
 m: \quad 2^{\Omega} \rightarrow [0, 1],
\end{equation}
satisfying the following conditions,
\begin{equation}
\label{eq_Massfunction2}
\begin{aligned}
 m(\emptyset) = 0, \text{ and }
 \sum\limits_{A \neq \emptyset | A \in 2^{\Omega}} m(A) = 1.
 \end{aligned}
\end{equation}
\end{definition}

In the DS evidence theory, $m$ can also be called a basic belief assignment (BBA).
If $m(A)$ is greater than zero, where $A \in 2^{\Omega}$, $A$ is called a focal element.
The value of $m(A)$ represents how strongly the evidence supports the proposition $A$.

\begin{definition}(Belief function)

Let $A$ be a proposition in the frame of discernment $\Omega$.
The belief function of proposition $A$, denoted as $Bel(A)$ is defined by
\begin{equation}\label{eq_belieffunction}
\begin{aligned}
Bel(A) = \sum\limits_{B \neq \emptyset | B \subseteq A} m(B).
\end{aligned}
\end{equation}
\end{definition}

\begin{definition}(Plausibility function)

Let $A$ be a proposition in the frame of discernment $\Omega$.
The plausibility function of proposition $A$, denoted as $Pl(A)$ is defined by
\begin{equation}\label{eq_plausibilityfunction}
\begin{aligned}
Pl(A) = \sum\limits_{B \neq \emptyset | B \cap A \neq \emptyset} m(B).
\end{aligned}
\end{equation}
\end{definition}

The belief function $Bel(A)$ and plausibility function $Pl(A)$ represent the lower and upper bound functions of the proposition $A$, respectively.

\section{The complex mass function}\label{complexmassfunction}

The $\Omega$ is defined as a frame of discernment which consists of a set of mutually exclusive and collective non-empty events, denoted as
\begin{equation}\label{eq_Frameofdiscernment1}
 \Omega = \{\psi_{1}, \psi_{2}, \ldots, \psi_{i}, \ldots, \psi_{N}\},
\end{equation}

The power set of $\Omega$ is represented by $2^{\Omega}$, denoted as
\begin{equation}\label{eq_Frameofdiscernment2}
\begin{aligned}
 2^{\Omega} = \{\emptyset, \{\psi_{1}\}, \{\psi_{2}\}, \ldots, \{\psi_{N}\}, \{\psi_{1}, \psi_{2}\}, \ldots, \{\psi_{1}, 
 \psi_{2}, \ldots, \psi_{i}\}, \ldots, \Omega\},
\end{aligned}
\end{equation}
where $\emptyset$ is an empty set.

When $A \in 2^{\Omega}$, $A$ is called a proposition or hypothesis.

\begin{definition}(Complex mass function)\label{def_Complexmassfunction}

A complex mass function $\mathds{M}$ in the frame of discernment $\Omega$ is modeled as a complex number, which is represented as a mapping from $2^{\Omega}$ to $\mathbb{C}$, defined by
\begin{equation}\label{eq_GMassfunction1}
 \mathds{M}: \quad 2^{\Omega} \rightarrow \mathbb{C},
\end{equation}
satisfying the following conditions,
\begin{equation}\label{eq_Massfunction2}
\begin{aligned}
 \mathds{M}(\emptyset) &= 0, \\
 \mathds{M}(A) &= \mathbf{m}(A) e^{i \theta(A)}, \quad A \neq \emptyset | A \in 2^{\Omega} \\
 \sum\limits_{A \neq \emptyset | A \in 2^{\Omega}} \mathds{M}(A) &= 1,
\end{aligned}
\end{equation}
where $i = \sqrt{-1}$; $\mathbf{m}(A) \in [0, 1]$ representing the magnitude of the complex mass function $\mathds{M}(A)$;
$\theta(A) \in [-\pi, \pi]$ denoting a phase term.
\end{definition}

In Eq.~(\ref{eq_Massfunction2}), $\mathds{M}(A)$ can also expressed in the ``rectangular'' form or ``Cartesian'' form, denoted by
\begin{equation}\label{eq_Massfunction3}
\mathds{M}(A) = x + yi, \quad A \neq \emptyset | A \in 2^{\Omega}
\end{equation}
with
\begin{equation}\label{eq_Massfunction4}
\sqrt{x^2+y^2} \in [0, 1].
\end{equation}

By using the Euler's relation, the magnitude and phase of the complex mass function $\mathds{M}(A)$ can be expressed as
\begin{equation}\label{eq_magnitudeandphase}
\mathbf{m}(A)=\sqrt{x^2+y^2}, \text{ and } \theta(A) = \arctan(\frac{y}{x}),
\end{equation}
where $x = \mathbf{m}(A) \cos(\theta(A))$ and $y = \mathbf{m}(A) \sin(\theta(A))$.

The square of the absolute value for $\mathds{M}(A)$ is defined by
\begin{equation}\label{eq_squaremagnitude}
|\mathds{M}(A)|^2=\mathds{M}(A)\mathds{\overline{M}}(A)=x^2+y^2,
\end{equation}
where $\mathds{\overline{M}}(A)$ is the complex conjugate of $\mathds{M}(A)$, such that $\mathds{\overline{M}}(A)=x - yi$.

These relationships can be then obtained as
\begin{equation}\label{eq_relationship}
\mathbf{m}(A)=|\mathds{M}(A)|, \text{ and } \theta(A) = \angle \mathds{M}(A),
\end{equation}
where if $\mathds{M}(A)$ is a real number (i.e., $y = 0$), then $\mathbf{m}(A) = |x|$.

The complex mass function $\mathds{M}$ modeled by the complex number can also be called a complex basic belief assignment (CBBA) in the evidence theory.

\begin{definition}(The focal element of CBBA)

Let $\mathds{M}$ be a complex mass function in the frame of discernment $\Omega$.
$A$ is called a focal element of CBBA $\mathds{M}$, when satisfying the following condition
\begin{equation}\label{eq_Focalelement}
|\mathds{M}(A)| > 0, \quad A \neq \emptyset | A \subseteq \Omega.
\end{equation}
\end{definition}

\begin{definition}(Commitment degree of CBBA)

Let $A$ be a hypothesis in the frame of discernment $\Omega$.
The commitment degree measure that is committed to $A$ for the CBBA $\mathds{M}$ is defined by
\begin{equation}\label{eq_Commitmentdegree}
\begin{aligned}
\mathds{C}om(A) &= \frac{|\mathds{M}(A)|}{\sum_{B \neq \emptyset | B \subseteq \Omega} |\mathds{M}(B)|}, \quad A \neq \emptyset | A \subseteq \Omega \\
&= \frac{\mathbf{m}(A)}{\sum_{B \neq \emptyset | B \subseteq \Omega} \mathbf{m}(B)}.
\end{aligned}
\end{equation}
\end{definition}

\section{Generalized belief and plausibility functions}\label{Generalizedbelieffunction}

\begin{definition}(Generalized belief function)

Given by the CBBA $\mathds{M}: 2 \rightarrow \mathbb{C}$.
The generalized belief function, denoted as $\mathds{B}el$ in the proposition $A \subseteq \Omega | A \neq \emptyset$ is defined by a mapping from $2^\Omega$ to [0, 1]:
\begin{equation}\label{eq_Gbelieffunction}
\mathds{B}el(A) = \sum\limits_{B \subseteq A} \mathds{C}om(B),
\end{equation}
where
\begin{equation}
\mathds{C}om(B) =
\frac{|\mathds{M}(B)|}{\sum\limits_{C \neq \emptyset | C \subseteq \Omega} |\mathds{M}(C)|}
=
\frac{\mathbf{m}(B)}{\sum\limits_{C \neq \emptyset | C \subseteq \Omega} \mathbf{m}(C)}.
\end{equation}
\end{definition}

When the CBBAs degrade into real numbers, i.e., traditional BBAs, the generalized belief function $\mathds{B}el$ will degrade into belief function $Bel$.

\begin{Theorem}
The generalized belief function $\mathds{B}el$ satisfies the following axioms:
\begin{enumerate}[\text{Axiom} 1:]
\item
$\mathds{B}el(\emptyset)= 0$,

\item
$\mathds{B}el(\Omega) = 1$,

\item
For $A_1, A_2, \ldots, A_m$ of subsets of $\Omega$,
\begin{equation}\label{eq_Axiom3}
\begin{aligned}
\mathds{B}el(A_1 \cup A_2 \ldots \cup A_m) \geq
&\sum_i \mathds{C}om(A_i) - \sum_{i<j} \mathds{C}om(A_i \cap A_j) \pm \\
&\ldots + (-1)^{m+1} \mathds{C}om(A_1 \cap A_2 \ldots \cap A_m) \\
&= \sum_{I \neq \emptyset | I \subset \{1,\ldots,m\}} (-1)^{|I|+1} \mathds{C}om(\bigcap_{i\in I} A_i).
\end{aligned}
\end{equation}
\end{enumerate}
\end{Theorem}

\begin{Theorem}\label{Theoremtransformation}
Let $\mathds{B}el: 2^\Omega \rightarrow [0, 1]$ be a generalized belief function given by the CBBA $\mathds{M}: 2 \rightarrow \mathbb{C}$.
For $A \subseteq \Omega | A \neq \emptyset$, the $\mathds{C}om(A)$ can be constructed by using a $M\ddot{o}bius$ transformation:
\begin{equation}\label{eq_transformation1}
\begin{aligned}
\mathds{C}om(A) = \sum_{B \subseteq A} (-1)^{|A-B|} \mathds{B}el(B),
\end{aligned}
\end{equation}
where $|A - B|$ is the cardinality of $A - B$.

\end{Theorem}

\begin{definition}(Generalized plausibility function)

Given by the CBBA $\mathds{M}: 2 \rightarrow \mathbb{C}$.
The generalized plausibility function, denoted as $\mathds{P}l(A)$ in the proposition $A \subseteq \Omega | A \neq \emptyset$ is defined by a mapping from $2^\Omega$ to [0, 1]:
\begin{equation}\label{eq_Gplausibilityfunction}
\begin{aligned}
\mathds{P}l(A) = 1-\mathds{B}el(\bar{A}) = 1-\sum\limits_{B \subseteq \bar{A}} \mathds{C}om(B) = \sum\limits_{B \cap A \neq \emptyset} \mathds{C}om(B).
\end{aligned}
\end{equation}
where $\bar{A}$ is the complement of $A$, such that $\bar{A} = \Omega - A$.
\end{definition}

When the CBBAs degrade into real numbers, i.e., traditional BBAs, the generalized plausibility function $\mathds{P}l$ will degrade into plausibility function $Pl$.

\section{Conclusions}\label{Conclusion}
In this paper, a complex mass function modeled by the complex number was proposed, called as a complex basic belief assignment.
Based on the definition of the complex mass function, we proposed new concepts of generalized belief and plausibility functions.
In particular, when the complex mass function was degenerated from complex numbers to real numbers, the generalized belief and plausibility functions degenerated into the traditional belief and plausibility functions, respectively.
In summary, this study is the first work to generalize the belief function in the framework of complex number.
It provides a promising way for uncertainty modelling and processing in decision theory.

\section*{Conflict of Interest}
The author states that there are no conflicts of interest.

\section*{Acknowledgment}
This research is supported by the Fundamental Research Funds for the Central Universities (No. XDJK2019C085) and Chongqing Overseas Scholars Innovation Program (No. cx2018077).

\clearpage
\normalsize
\bibliographystyle{elsarticle-num}

\end{document}